\documentclass[sigconf]{acmart}
\setcopyright{none}
\renewcommand\footnotetextcopyrightpermission[1]{}
\AtBeginDocument{%
  }
\usepackage{algorithm}
\usepackage{algorithmic}
\usepackage{amsmath} % 用于数学符号

\usepackage{booktabs} 
\usepackage{graphicx}
\usepackage{multirow}  
\usepackage[table]{xcolor}
\usepackage{siunitx}
\usepackage{hyperref}
\usepackage{threeparttable}

\begin{document}

\title{SAGE: Sustainable Agent-Guided Expert-tuning for Culturally Attuned Translation in Low-Resource Southeast Asia}

% \author{Anonymous Author}
% MERIT-LMIC: Multilingual Expert-Reward Informed Tuning for Low-Resource Southeast-Asian Translation and Community Dialogues

\author{Zhixiang Lu}\authornote{Equal contribution.}
\email{zhixiang.lu22@student.xjtlu.edu.cn}
\affiliation{
  \institution{Xi'an Jiaotong-Liverpool University}
  \city{Suzhou} % 建议补充城市，虽非必须
  \country{China}
}

\author{Chong Zhang}\authornotemark[1]
\email{c.zhang118@liverpool.ac.uk}
\affiliation{
  \institution{Xi'an Jiaotong-Liverpool University}
  \city{Suzhou}
  \country{China}
}

\author{Yulong Li}\authornotemark[1]
\email{yulong.li19@student.xjtlu.edu.cn}
\affiliation{
  \institution{Xi'an Jiaotong-Liverpool University}
  \city{Suzhou}
  \country{China}
}

\author{Angelos Stefanidis}
\email{angelos.stefanidis@xjtlu.edu.cn}
\affiliation{
  \institution{Xi'an Jiaotong-Liverpool University}
  \city{Suzhou}
  \country{China}
}

\author{Anh Nguyen}
\email{anh.nguyen@liverpool.ac.uk}
\affiliation{
  \institution{University of Liverpool}
  \city{Liverpool}
  \country{United Kingdom}
}

\author{Imran Razzak}\authornotemark[2]
\email{imran.razzak@mbzuai.ac.ae}
\affiliation{
  \institution{Mohamed bin Zayed University of Artificial Intelligence}
  \city{Abu Dhabi}
  \country{United Arab Emirates}
}

\author{Jionglong Su}\authornotemark[2]
\email{jionglong.su@xjtlu.edu.cn}
\affiliation{
  \institution{Xi'an Jiaotong-Liverpool University}
  \city{Suzhou}
  \country{China}
}

\author{Zhengyong Jiang}
\authornote{Corresponding author.}
\email{zhengyong.jiang02@xjtlu.edu.cn}
\affiliation{
  \institution{Xi'an Jiaotong-Liverpool University}
  \city{Suzhou}
  \country{China}
}

\renewcommand{\shortauthors}{Zhixiang Lu et al.}
\begin{abstract}
The vision of an inclusive World Wide Web is impeded by a severe linguistic divide, particularly for communities in low-resource regions of Southeast Asia. While large language models (LLMs) offer a potential solution for translation, their deployment in data-poor contexts faces a dual challenge: the scarcity of high-quality, culturally relevant data and the prohibitive energy costs of training on massive, noisy web corpora. To resolve the tension between digital inclusion and environmental sustainability, we introduce Sustainable Agent-Guided Expert-tuning (SAGE). This framework pioneers an energy-aware paradigm that prioritizes the "right data" over "big data". Instead of carbon-intensive training on unfiltered datasets, SAGE employs a reinforcement learning (RL) agent, optimized via Group Relative Policy Optimization (GRPO), to autonomously curate a compact training set. The agent utilizes a semantic reward signal derived from a small, expert-constructed set of community dialogues to filter out noise and cultural misalignment. We then efficiently fine-tune open-source LLMs on this curated data using Low-Rank Adaptation (LoRA). We applied SAGE to translation tasks between English and seven low-resource languages (LRLs) in Southeast Asia. Our approach establishes new state-of-the-art performance on BLEU-4 and COMET-22 metrics, effectively capturing local linguistic nuances. Crucially, SAGE surpasses baselines trained on full datasets while reducing data usage by 97.1\% and training energy consumption by 95.2\%. By delivering high-performance models with a minimal environmental footprint, SAGE offers a scalable and responsible pathway to bridge the digital divide in the Global South.

\end{abstract}
%    <concept>
%        <concept_id>10010147.10010178.10010179.10010180</concept_id>
%        <concept_desc>Computing methodologies~Machine translation</concept_desc>
%        <concept_significance>500</concept_significance>
%        </concept>
%    <concept>
%        <concept_id>10010147.10010178.10010179.10010186</concept_id>
%        <concept_desc>Computing methodologies~Language resources</concept_desc>
%        <concept_significance>300</concept_significance>
%        </concept>
%  </ccs2012>
% \ccsdesc[500]{Computing methodologies~Machine translation}
% \ccsdesc[300]{Computing methodologies~Language resources}
\maketitle

\section{Introduction}

The proliferation of Large Language Models (LLMs) has catalyzed a revolution in automated communication, with Neural Machine Translation (NMT) systems achieving remarkable fluency and accuracy across high-resource language pairs. Architectures like the Transformer \cite{vaswani2017attention} have become foundational, enabling seamless interaction and information exchange for speakers of languages such as English, Spanish, and French. However, this progress has not been universally distributed. A stark digital and linguistic divide persists, as the performance of these data-hungry models remains profoundly suboptimal for the vast majority of the world's approximately 7,000 low-resource languages (LRLs). This disparity is primarily due to the lack of large-scale, high-quality parallel corpora needed for training. As a result, entire populations, especially in Low and Middle-Income Countries (LMICs), are excluded from the advantages of modern AI, limiting their access to global information and digital services. The current trend in NMT development unintentionally exacerbates existing inequalities. By prioritizing methods that rely on large datasets, it creates a positive feedback loop for high-resource languages. In contrast, LRLs are trapped in a negative cycle characterized by data scarcity, poor model performance, and low user adoption. To tackle this systemic issue, a fundamental shift in methodology is essential.

This challenge is most acute and socially consequential in the domain of \textit{community dialogues}: the informal, context-rich, and culturally nuanced conversations that are the bedrock of civic life. These dialogues cover critical topics such as public health advisories, local commerce, and educational support. Standard MT systems, typically trained on formal corpora like news articles or parliamentary records, consistently fail to capture the subtleties of this domain. They struggle with the ambiguity inherent in short texts, colloquialisms, code-switching, and culturally specific idioms, which are hallmarks of community interaction. The consequences of such failures are not merely linguistic; they are social. Inaccurate or culturally insensitive translations can erode trust, disengage culturally and linguistically diverse (CALD) communities from vital public health campaigns, and obstruct access to educational materials for students who rely on mobile devices for learning. In this context, translation quality is not an abstract technical metric but a direct determinant of social impact. A model that produces grammatically correct but contextually inappropriate translations can do more harm than good.

The predominant strategy for addressing low-resource scenarios has been a "\textit{more data}" paradigm, focused on augmenting limited datasets, primarily through back-translation from large monolingual corpora to generate synthetic parallel data. While this can yield improvements, it often amplifies the noise inherent in the vast, unfiltered web data from which monolingual corpora are sourced \cite{haddow2022survey,yin2024lexmatcher}. Critically, simply increasing the volume of generic, out-of-domain data does not guarantee improved performance on the specialized task of translating community dialogues. Indeed, fine-tuning an LLM on a massive corpus that is 99\% generic web text may degrade its ability to handle the rare but crucial linguistic patterns of the target domain. This suggests that a paradigm shift is necessary: from a focus on "more data" to a focus on the "right data". The central technical challenge thus becomes the autonomous and scalable curation of a compact, high-potency, in-domain dataset from a much larger, noisy corpus.

\begin{figure}[t]
    \centering
    \includegraphics[width=\linewidth]{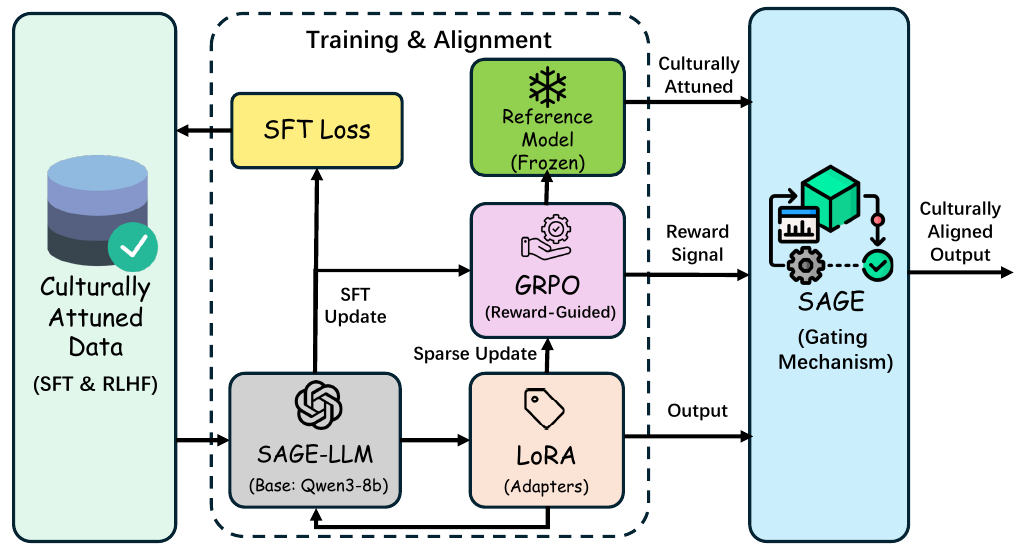}
    \caption{
        Architectural overview of the SAGE framework.
    }
    \label{fig:sage_simple}
    \vspace{-2em}
\end{figure}

To address this challenge, we introduce SAGE: \textbf{S}ustainable \textbf{A}gent-\textbf{G}uided \textbf{E}xpert-tuning (see \ref{fig:sage_simple}). This framework operationalizes the "right data" philosophy by pioneering an expert-reward informed tuning paradigm. Instead of fine-tuning on a noisy, unfiltered corpus, SAGE first employs a Reinforcement Learning (RL) agent to autonomously curate a small, high-quality training set. The agent's policy is trained using Group Relative Policy Optimization (GRPO) \cite{shao2024deepseekmath}, a highly efficient, critic-free RL algorithm. The novelty lies in our reward signal: the semantic similarity between a candidate translation and a small, "golden" reference set of expert-translated community dialogues. This mechanism distills expert domain knowledge and cultural attunement directly into the data selection process. Subsequently, a powerful open-source LLM is efficiently fine-tuned on this curated dataset using Low-Rank Adaptation (LoRA) \cite{hu2021lora}. Applied to a challenging multilingual task involving English and seven low-resource Southeast Asian languages (Burmese, Bengali, Filipino, Hindi, Khmer, Lao, and Vietnamese), SAGE sets a new state-of-the-art (SOTA) on key metrics like BLEU-4 and COMET-22 \cite{rei-etal-2022-comet}, surpassing baselines trained on the entire unfiltered dataset while using substantially less data.
\noindent Our contributions are threefold:
\begin{itemize}
\item We propose SAGE, a novel framework that pioneers the use of RL with an expert-defined reward for autonomous, quality-driven data curation in low-resource MT.
\item We introduce a new application of GRPO, leveraging its efficiency for the upstream data selection task, guided by a semantic-similarity signal that encodes expert domain knowledge.
\item We demonstrate SOTA translation performance across seven low-resource Southeast Asian languages, validating that our "right data" approach is more effective and resource-efficient than the conventional "more data" paradigm.
\end{itemize}

\section{Related Work}

\subsection{Machine Translation for LRLs}
NMT has established itself as the dominant paradigm in translation tasks. Formally, given a source sentence $\mathbf{x} = (x_1, \dots, x_M)$ and a target sentence $\mathbf{y} = (y_1, \dots, y_N)$, NMT models aim to maximize the log-likelihood of the conditional probability:
\begin{equation}
    \mathcal{L}_{\text{NMT}}(\theta) = \sum_{(\mathbf{x}, \mathbf{y}) \in \mathcal{D}} \sum_{t=1}^{N} \log P(y_t | y_{<t}, \mathbf{x}; \theta)
\end{equation}
where $\theta$ represents the model parameters and $\mathcal{D}$ is the parallel corpus. While Transformer-based architectures \cite{vaswani2017attention} have achieved remarkable success in high-resource scenarios \cite{wu2016google}, they face significant performance degradation in LRLs due to the sparsity of $\mathcal{D}$ \cite{koehn2017neural, neubig2018neural}.

To mitigate this, traditional approaches leverage transfer learning \cite{johnson2017google}, back-translation \cite{sennrich2016edinburgh}, and unsupervised MT \cite{lample2018unsupervised}. Recently, LLMs such as mBERT \cite{devlin2019bert}, XLM-R \cite{conneau2020unsupervised}, and NLLB \cite{team2022no} have demonstrated strong zero-shot capabilities. However, general-purpose LLMs often lack the granularity required for domain-specific tasks in LMICs, necessitating targeted fine-tuning strategies.
\begin{figure*}[t]
  \centering
  \includegraphics[width=\textwidth]{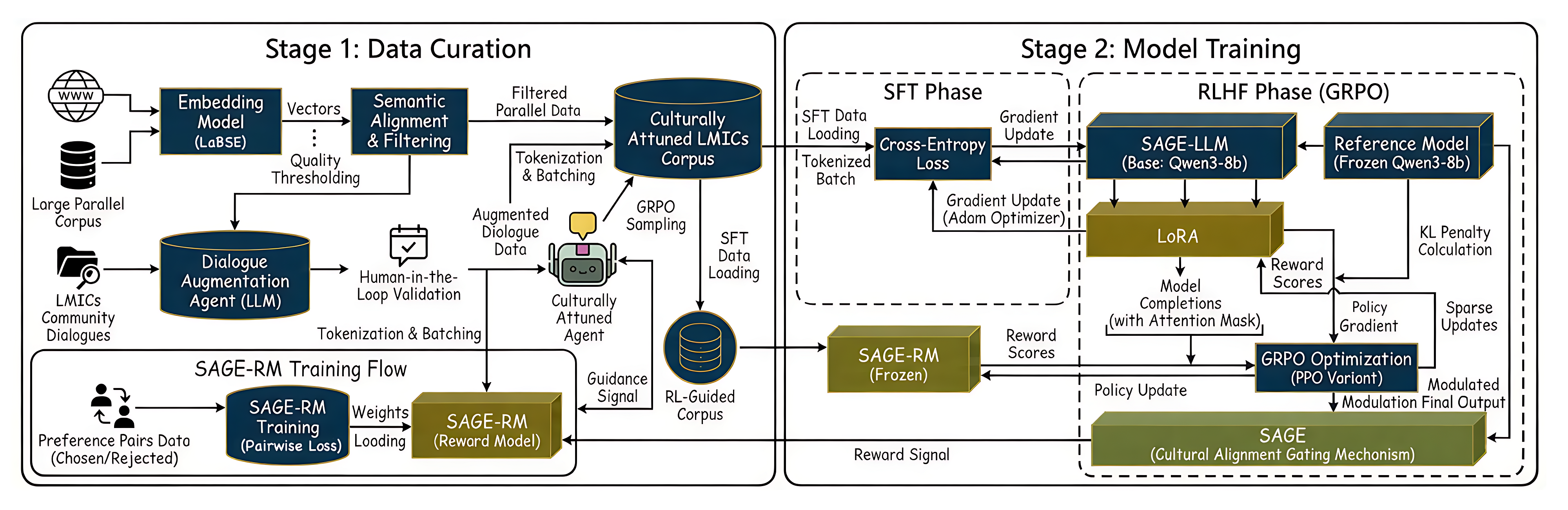}
  \vspace{-2em}
  \caption{The SAGE training and alignment pipeline. Stage 1 employs a GRPO-optimized RL agent to curate a subset $\mathcal{D}_{\text{cur}}$ from a noisy pool $\mathcal{D}_{\text{noisy}}$, guided by semantic proximity to a small expert reference $\mathcal{D}_{\text{exp}}$. Stage 2 utilizes LoRA to efficiently fine-tune the LLM on $\mathcal{D}_{\text{cur}}$, minimizing computational overhead while maximizing cultural alignment.}
  \label{fig:framework}
  \vspace{-1em}
\end{figure*}
\subsection{Data Curation and Quality Estimation}
The efficacy of NMT is heavily contingent on data quality \cite{van2017dynamic}. In low-resource settings, available corpora are often plagued by noise and domain misalignment \cite{khayrallah2018impact, lu2025}. Existing filtering techniques typically employ heuristic scoring functions $S(\mathbf{x}, \mathbf{y})$ based on length ratios or language identification probabilities, discarding pairs where $S(\mathbf{x}, \mathbf{y}) < \tau$ \cite{koehn2005europarl, ott2018filtering}. More advanced methods utilize dual cross-entropy loss or quality estimation models \cite{fan2020beyond}. While active learning strategies \cite{settles2009active} attempt to optimize sample efficiency, they remain dependent on expensive human annotation. Unlike these rule-based or human-in-the-loop approaches~\cite{toneva2019empirical, settles2009active, ren2018learning,whang2023data}, our work introduces an autonomous reinforcement learning framework~\cite{yoon2020data} guided by semantic ``expert-reward'' signals~\cite{wu2024self}, shifting the paradigm towards intelligent, goal-oriented data curation \cite{garg2023rlaif}.

\subsection{Reinforcement Learning for Data Selection}
RL has been widely adopted in NLP to optimize non-differentiable metrics~\cite{yoon2020data}. The data selection process can be formulated as a Markov Decision Process (MDP) \cite{MDR25}, where the agent's policy $\pi_\phi(a|s)$ selects data subsets to maximize a downstream reward $R$. The objective is to maximize the expected reward:
\begin{equation}
    J(\phi) = \mathbb{E}_{\tau \sim \pi_\phi} [R(\tau)]
\end{equation}
Previous works have applied RL to curriculum learning and instance weighting \cite{gong2020reinforcement, zhang2021reinforcement}. However, stability remains a challenge in RL optimization. Our SAGE framework employs GRPO, a method that stabilizes training by normalizing advantages within group samples. This allows our agent to effectively learn a policy that aligns the training data distribution with the semantic and cultural nuances of expert-curated community dialogues, representing a novel application of GRPO in data curation.

\subsection{Parameter-Efficient Fine-Tuning}
Full fine-tuning of LLMs is computationally prohibitive for many applications in the Global South. Parameter-Efficient Fine-Tuning (PEFT) addresses this by updating only a small subset of parameters. LoRA \cite{hu2021lora}, a prominent PEFT method, hypothesizes that the change in weights $\Delta W$ has a low intrinsic rank. For a pre-trained weight matrix $W_0 \in \mathbb{R}^{d \times k}$, LoRA decomposes the update as:
\begin{equation}
    W = W_0 + \Delta W = W_0 + B A
\end{equation}
where $B \in \mathbb{R}^{d \times r}$ and $A \in \mathbb{R}^{r \times k}$ are trainable matrices with rank $r \ll \min(d, k)$. This technique drastically reduces memory requirements while maintaining performance comparable to full fine-tuning \cite{houlsby2019parameter, li2021prefix, liu2021p}. In SAGE, we leverage LoRA to efficiently adapt LLMs to our RL-curated data, ensuring the system remains deployable in resource-constrained environments typical of LMICs.

\subsection{Translation for Community Empowerment}
Translating for LMICs extends beyond linguistic accuracy to cultural and contextual appropriateness. Community dialogues frequently exhibit code-switching \cite{huzaifah-etal-2024-evaluating}, non-standard orthography, and localized idioms \cite{donthi2025improving} that are absent in standard benchmarks \cite{guzman2019challenges}. Furthermore, ethical AI deployment in these regions requires systems that empower users rather than merely extracting data \cite{bird-etal-2023-ethical, zhang-etal-2024-mc2}. Addressing the lack of culturally sensitive NLP systems \cite{blodgett2020language, hershcovich-etal-2022-challenges}, our framework explicitly models these factors via the reward mechanism, ensuring translations preserve the semantic integrity and cultural intent vital for community engagement.

\section{Methodology}
\label{sec:methodology}

Our SAGE framework addresses the misalignment between general-purpose LLMs and community-specific linguistic nuances in low-resource settings. We formulate the problem as a two-stage pipeline: (1) \textit{Expert-Informed Data Curation}, where a RL agent, optimized via GRPO, autonomously selects high-leverage training samples; and (2) \textit{Parameter-Efficient Fine-Tuning}, where the selected data minimizes the domain shift using LoRA. The overall architecture is illustrated in Figure~\ref{fig:framework}.

\subsection{Problem Formulation}
Let $\mathcal{D}_{\text{noisy}} = \{(x_i, y_i)\}_{i=1}^{N}$ denote a large-scale, general-domain parallel corpus, where $x_i$ and $y_i$ represent source and target sentences, respectively. We assume the distribution of $\mathcal{D}_{\text{noisy}}$ diverges from the target community domain. Conversely, we possess a small, expert-verified reference set $\mathcal{D}_{\text{exp}} = \{(x'_j, y'_j)\}_{j=1}^{M}$, where $M \ll N$.

Our objective is to learn a selection policy $\pi_\theta$ that identifies a subset $\mathcal{D}_{\text{cur}} \subset \mathcal{D}_{\text{noisy}}$ with cardinality $|\mathcal{D}_{\text{cur}}| = K$, such that the distributional distance between $\mathcal{D}_{\text{cur}}$ and $\mathcal{D}_{\text{exp}}$ is minimized. Subsequently, we optimize a translation model $\Phi$ on $\mathcal{D}_{\text{cur}}$ to maximize the likelihood of the target domain translations.
\begin{algorithm}[t]
\caption{Expert-Guided Corpus Curation Strategy}
\label{alg:data_curation}
\begin{algorithmic}[1]
\REQUIRE Noisy Corpus $\mathcal{D}_{\text{noisy}}$, Expert Reference $\mathcal{D}_{\text{exp}}$, Budget $K$
\REQUIRE Pre-trained Policy $\pi_{\theta^\star}$, Encoder $\mathbf{E}$
\STATE Initialize $\mathcal{D}_{\text{cur}} \leftarrow \emptyset$, $\mathcal{P} \leftarrow \mathcal{D}_{\text{noisy}}$
\STATE \textbf{Pre-compute} expert embeddings: $\mathbf{V}_{\text{exp}} \leftarrow \{\mathbf{E}(y') \mid (x', y') \in \mathcal{D}_{\text{exp}}\}$
\WHILE{$|\mathcal{D}_{\text{cur}}| < K$}
    \STATE Evaluate candidates in $\mathcal{P}$ using Policy $\pi_{\theta^\star}$:
    \STATE $\forall (x, y) \in \mathcal{P}, \quad \text{score}(x, y) \leftarrow \pi_{\theta^\star}( (x, y) \mid \mathcal{D}_{\text{cur}} )$
    \STATE Select top candidate:
    \STATE $(x^*, y^*) \leftarrow \arg\max_{(x,y) \in \mathcal{P}} \text{score}(x, y)$
    \STATE Update sets:
    \STATE $\mathcal{D}_{\text{cur}} \leftarrow \mathcal{D}_{\text{cur}} \cup \{(x^*, y^*)\}$
    \STATE $\mathcal{P} \leftarrow \mathcal{P} \setminus \{(x^*, y^*)\}$
\ENDWHILE
\RETURN $\mathcal{D}_{\text{cur}}$
\end{algorithmic}
\end{algorithm}
\subsection{RL-Guided Data Curation}
% We model the data curation process as a Markov Decision Process (MDP) defined by the tuple $\langle \mathcal{S}, \mathcal{A}, \mathcal{P}, \mathcal{R} \rangle$.

\subsubsection{MDP Definitions}
\begin{itemize}
    \item \textbf{State Space ($\mathcal{S}$):} The state $s_t$ represents the current curated subset at step $t$, i.e., $s_t = \mathcal{D}_{\text{cur}}^{(t)}$. The initial state is $s_0 = \emptyset$.
    \item \textbf{Action Space ($\mathcal{A}$):} An action $a_t$ consists of selecting a candidate pair $(x_k, y_k)$ from the remaining pool $\mathcal{U}_t = \mathcal{D}_{\text{noisy}} \setminus \mathcal{D}_{\text{cur}}^{(t)}$.
    \item \textbf{Reward Function ($\mathcal{R}$):} To guide the agent towards community-aligned data, we employ a dense reward signal based on semantic embedding similarity. Let $\mathbf{E}(\cdot)$ denote a pre-trained sentence encoder (LaBSE \cite{feng-etal-2022-language}). The reward for selecting action $a_t = (x, y)$ is defined as the mean cosine similarity against the expert reference:
    \begin{equation}
        r(s_t, a_t) = \frac{1}{M} \sum_{(x', y') \in \mathcal{D}_{\text{exp}}} \frac{\mathbf{E}(y)^\top \mathbf{E}(y')}{\|\mathbf{E}(y)\| \|\mathbf{E}(y')\|}
    \end{equation}
    This formulation explicitly encourages the selection of instances that share semantic and stylistic features with high-quality community dialogues.
\end{itemize}

\subsubsection{GRPO}
Standard policy gradient methods often suffer from high variance in reward estimation. We employ GRPO \cite{yuan2023group} to stabilize learning by leveraging relative preferences between trajectories.

Specifically, for a given input state, we sample a group of trajectories (selection sequences) $\{\tau_1, \tau_2, \dots, \tau_G\}$. We define a pairwise preference probability using the Bradley-Terry model. The objective is to maximize the expected log-likelihood of the preferred trajectory $\tau_w$ over a less optimal trajectory $\tau_l$:
\begin{equation}
    \mathcal{L}_{\text{GRPO}}(\theta) = -\mathbb{E}_{(\tau_w, \tau_l) \sim \pi_\theta} \left[ \log \sigma \left( \beta \left( \sum_{t} r_t^{(w)} - \sum_{t} r_t^{(l)} \right) \right) \right]
\end{equation}
where $\sigma$ is the sigmoid function and $\beta$ is a temperature parameter. This approach allows the agent to learn robust selection criteria based on relative quality rather than absolute, noisy reward values.

\subsection{Parameter-Efficient Fine-Tuning}
To adapt the LLM $\Phi$ to the curated dataset $\mathcal{D}_{\text{cur}}$ under resource constraints, we utilize LoRA \cite{hu2021lora}. 

For a pre-trained weight matrix $W_0 \in \mathbb{R}^{d \times k}$ in the transformer layers, we constrain the update $\Delta W$ by representing it as the product of two low-rank matrices $B \in \mathbb{R}^{d \times r}$ and $A \in \mathbb{R}^{r \times k}$, where $r \ll \min(d, k)$. The forward pass is formalized as:
\begin{equation}
    h = W_0 x + \Delta W x = W_0 x + \frac{\alpha}{r} BA x
\end{equation}
where $\alpha$ is a scaling factor. During training, $W_0$ is frozen, and only $A$ and $B$ are optimized. The final training objective minimizes the negative log-likelihood over the curated subset:
\begin{equation}
    \mathcal{L}_{\text{FT}}(\Phi) = - \sum_{(x, y) \in \mathcal{D}_{\text{cur}}} \log P_{\Phi}(y \mid x; W_0, A, B)
\end{equation}
This ensures that the model adapts to the specific linguistic properties of the community data while maintaining the generalization capabilities of the base model.

\begin{table*}[t!]
\centering
% \vspace{10pt}
\caption{Comparison for English-to-Southeast-Asian translation in LMICs. The best result is in \textbf{bold}, the second best is \underline{underlined}. Avg. Tok. denotes estimated average inference token consumption per sample.}
\label{tab:main_results}
\vspace{-5pt}
\small
\setlength{\tabcolsep}{4pt}
\sisetup{
    detect-weight,
    mode=text,
    table-format=2.2,
}

\begin{tabular}{@{}l *{7}{S} @{\hspace{6pt}} *{7}{S} @{\hspace{6pt}} S@{}} 
\toprule
\multicolumn{1}{c}{\multirow{2}{*}{\textbf{Model}}} & \multicolumn{7}{c}{\textbf{BLEU-4} $\uparrow$} & \multicolumn{7}{c}{\textbf{COMET-22} $\uparrow$} & {\multirow{2}{*}{\textbf{Avg. Tok.} $\downarrow$}} \\
\cmidrule(lr){2-8} \cmidrule(lr){9-15}
 & {\textbf{bn}} & {\textbf{fil}} & {\textbf{hi}} & {\textbf{km}} & {\textbf{lo}} & {\textbf{my}} & {\textbf{vi}} & {\textbf{bn}} & {\textbf{fil}} & {\textbf{hi}} & {\textbf{km}} & {\textbf{lo}} & {\textbf{my}} & {\textbf{vi}} & {} \\
\midrule
\rowcolor{gray!10}
\multicolumn{16}{l}{\textbf{\textit{Closed-Source Models}}} \\
\midrule
GPT-4o        & 40.15          & 45.88          & 42.50          & 35.10          & 28.33          & 31.05          & 45.13          & 83.50          & 84.10          & 84.05          & 81.90          & 79.25          & 80.11          & 85.55          & 92.50 \\
Claude-3.5 Sonnet    & 41.25          & 45.15          & 42.18          & 34.95          & 32.12          & \textbf{33.85} & 45.24          & 84.10          & 83.95          & 83.99          & 81.75          & 80.14          & \textbf{82.45} & 85.79          & 95.10 \\
Grok-3        & 41.50          & 46.10          & 43.85          & 36.20          & 33.55          & 32.90          & 46.15          & 84.10          & 84.90          & 84.95          & 82.88          & 81.50          & 81.75          & 86.10          & 93.40 \\
Gemini-2.5 pro    & 38.20          & 42.55          & 40.13          & 32.80          & 26.57          & 29.88          & 42.01          & 82.15          & 82.50          & 82.88          & 80.50          & 78.89          & 79.50          & 84.60          & 90.20 \\
\midrule
\rowcolor{gray!10}
\multicolumn{16}{l}{\textbf{\textit{Open-Source Models}}} \\
\midrule
DeepSeek-v3   & 41.85          & 46.50          & 44.10          & 36.95          & 33.10          & 33.50          & 46.80          & 84.55          & 85.05          & 85.15          & 83.10          & 81.85          & 81.60          & 86.05          & 75.60 \\
Gemma-3-9B     & 37.10          & 43.10          & 38.90          & 31.85          & 29.13          & 27.05          & 44.25          & 83.30          & 83.65          & 83.60          & 81.33          & 80.10          & 79.92          & 85.30          & 82.30 \\
Qwen-3-8B      & 37.50          & 43.85          & 39.55          & 32.15          & 29.40          & 27.50          & 44.88          & 83.45          & 83.90          & 83.80          & 81.50          & 80.25          & 80.10          & 85.65          & 62.10 \\
Llama-3.1-8B   & 36.80          & 43.55          & 39.10          & 31.50          & 28.81          & 26.15          & 44.50          & 83.15          & 83.80          & 83.50          & 81.10          & 79.95          & 79.80          & 85.45          & 78.40 \\
NLLB-200-3.3B & 22.40          & 25.55          & 24.18          & 20.15          & 16.13          & 21.15          & 25.80          & 76.50          & 77.85          & 77.01          & 76.10          & 75.83          & 77.05          & 78.81          & \textbf{45.20} \\
M2M-100-1.2B  & 3.11           & 2.25           & 2.90           & 1.88           & 0.05           & 0.02           & 3.41           & 61.88          & 60.15          & 61.05          & 59.80          & 56.01          & 56.13          & 62.40          & \underline{46.80} \\
\midrule
\rowcolor{gray!10}
\multicolumn{16}{l}{\textbf{\textit{Our Method}}} \\
\midrule
\textbf{SAGE} (Qwen-3-8B)   & \textbf{47.15} & \textbf{48.55} & \textbf{48.80} & \textbf{41.50} & \textbf{37.10} & 33.15          & \textbf{48.95} & \textbf{86.30} & \textbf{86.75} & \textbf{86.90} & \textbf{84.55} & \textbf{83.90} & 82.15          & \textbf{86.95} & 60.50 \\
\textbf{SAGE} (Llama-3.1-8B) & \underline{46.90} & \underline{48.20} & \underline{48.55} & 40.80          & 36.55          & 32.50          & \underline{48.60} & \underline{86.10} & \underline{86.50} & \underline{86.75} & \underline{84.20} & \underline{83.65} & 81.90          & \underline{86.70} & 76.20 \\
\textbf{SAGE} (Gemma-3-9B)  & 46.75          & 47.90          & 48.30          & \underline{41.15} & \underline{36.88} & \underline{33.25} & 48.45          & 85.95          & 86.40          & 86.60          & 84.05          & 83.50          & \underline{82.20} & 86.65          & 80.10 \\
\bottomrule
\end{tabular}
\vspace{-1em}
\end{table*}

\section{Experiments}
\label{sec:experiments}

\subsection{Experimental Setup}

\subsubsection{Datasets and Benchmarks}
We evaluate SAGE on a comprehensive low-resource benchmark covering seven linguistically diverse Southeast Asian languages. The data setup comprises three distinct strata:
\begin{enumerate}

    \item \textbf{Noisy Pre-training Corpus ($\mathcal{D}_{\text{noisy}}$):} A large-scale amalgamation of web-scraped data sourced from CCMatrix~\cite{schwenk2021ccmatrix}, CCAligned~\cite{el2020ccaligned}, and ParaCrawl~\cite{banon2020paracrawl}, totaling over 50M sentence pairs. This dataset represents the typical ``high-quantity, low-quality'' regime found in wild data curation scenarios, characterized by significant semantic noise and misalignment.
    
    \item \textbf{ALT Dataset ($\mathcal{D}_{\text{eval}}$):} The Asian Language Treebank \cite{riza2016asian}, a high-quality, multi-way parallel corpus covering English and several low-resource Asian languages (e.g., Filipino, Khmer, Lao). Unlike the noisy web corpus, ALT serves as a clean, human-curated benchmark to rigorously evaluate the model's robustness in low-resource adaptation settings.

    \item \textbf{Noisy Pre-training Corpus ($\mathcal{D}_{\text{noisy}}$):} A large-scale amalgamation of web-scraped data from CCMatrix, CCAligned, and ParaCrawl, totaling over 50M sentence pairs. This represents the typical "high-quantity, low-quality" data available in the wild.
    \item \textbf{Expert Reference Set ($\mathcal{D}_{\text{expert}}$):} Our core contribution, consisting of 2,000 high-quality parallel pairs per language. Curated by professional translators, this set focuses strictly on high-value community domains (healthcare, civic engagement).
    \item \textbf{Test Set ($\mathcal{D}_{\text{test}}$):} A held-out set of 500 sentences per language, strictly separated from $\mathcal{D}_{\text{expert}}$ to prevent data leakage.
\end{enumerate}

\subsubsection{Evaluation Metrics}
To provide a holistic assessment of translation quality, we employ a dual-metric strategy that balances surface-level lexical precision with deep semantic fidelity.

\noindent \textbf{Lexical Precision (BLEU-4).} 
We report BLEU-4 \cite{papineni2002bleu}, the de facto standard in machine translation research. BLEU computes the geometric mean of n-gram precision ($n=1\dots4$) between the hypothesis and reference. Despite its well-documented inability to capture synonymous phrasing or semantic shifts \cite{mathur2020tangled}, we include it to ensure strict comparability with prior literature and to evaluate the model's ability to generate exact lexical matches for domain-specific terminology. To guarantee reproducibility, we utilize the standardized SacreBLEU implementation \cite{post2018call}. It is computed as the geometric mean of modified precisions $p_n$, scaled by a Brevity Penalty (BP) to penalize short generations:
    \begin{equation}
    \begin{aligned}
        \text{BLEU-4} &= \text{BP} \cdot \exp \left( \sum_{n=1}^{4} w_n \log p_n \right), \\
        \text{BP} &= \begin{cases} 
        1 & \text{if } c > r \\
        e^{1 - r/c} & \text{if } c \leq r 
        \end{cases}
    \end{aligned}
    \end{equation}
    where $c$ is the candidate length, $r$ is the reference length, and $w_n = 1/4$ are uniform weights. The precision $p_n$ is computed using clipped $n$-gram counts to prevent over-generation rewards.

\noindent \textbf{Semantic Fidelity (COMET-22).} 
To address the limitations of n-gram metrics, we employ COMET-22 \cite{rei2022comet}, which leverages a pre-trained cross-lingual encoder (XLM-R \cite{conneau2020unsupervised}) to map inputs into a continuous semantic space. Let $s$, $h$, and $r$ denote the source, hypothesis, and reference sequences, respectively. The sentence embedding $\mathbf{e} \in \mathbb{R}^d$ is derived via \textit{Layer-wise Scalar Mixing}, which aggregates representations from all $L$ transformer layers. For any input sequence $x \in \{s, h, r\}$, the embedding is computed as:
\begin{equation}
    \mathbf{e}_x = \Omega(x) = \sum_{l=0}^{L} \frac{\exp(\alpha_l)}{\sum_{k=0}^{L} \exp(\alpha_k)} \cdot \text{POOL}(\mathbf{H}^l_x)
\end{equation}
where $\mathbf{H}^l_x$ is the hidden state of the $l$-th layer, $\alpha_l$ are trainable scalar weights, and $\text{POOL}$ denotes the extraction of the \texttt{[CLS]} token. 

To capture fine-grained semantic discrepancies, the model constructs a joint interaction feature space. We define a difference function $\mathcal{K}(\mathbf{u}, \mathbf{v})$ that concatenates the vectors, their element-wise (Hadamard) product $\odot$, and their absolute difference $|\cdot|$:
\begin{align}
    \mathcal{K}(\mathbf{u}, \mathbf{v}) &= [\mathbf{u}; \mathbf{v}; \mathbf{u} \odot \mathbf{v}; |\mathbf{u} - \mathbf{v}|] \\
    \mathbf{x}_{\text{fuse}} &= [\mathcal{K}(\mathbf{e}_h, \mathbf{e}_s); \mathcal{K}(\mathbf{e}_h, \mathbf{e}_r)]
\end{align}
The final quality score $\hat{y}$ is then predicted via a feed-forward regressor over the fused features $\mathbf{x}_{\text{fuse}}$:
\begin{equation}
    \hat{y} = \text{MLP}(\mathbf{x}_{\text{fuse}})
\end{equation}
% This formulation allows COMET to penalize semantic hallucinations by explicitly modeling the distance and interaction between the hypothesis and the expert reference in the latent space.

\subsubsection{Implementation Details}
All experiments were conducted on a node equipped with 8 $\times$ NVIDIA A100-80GB GPUs. For the SAGE framework, we employed the Qwen-3-8B as the base model. The RL agent utilized a lightweight BERT-based reward model. Fine-tuning was performed using LoRA with rank $r=64$, alpha $\alpha=16$, and a learning rate of $2e-4$.

\subsection{Comparative Analysis}
\definecolor{forestgreen}{RGB}{34, 139, 34}
\definecolor{rowgray}{gray}{0.95}

\newcommand{\reduction}[1]{\textcolor{forestgreen}{\scriptsize{(-#1\%)}}}

\begin{table*}[t]
\renewcommand{\arraystretch}{1.1}
\centering
\begin{threeparttable} % <--- 1. 开始 threeparttable 环境
    \caption{Ablation study of the SAGE framework on the Qwen-3-8B model. We analyze the impact of removing key components (w/o) on BLEU-4 scores across 7 languages. The rightmost column demonstrates the environmental efficiency of SAGE compared to the baseline.}
    \label{tab:ablation_final_complete}
    \small
    \setlength{\tabcolsep}{3.2pt}
    
    \begin{tabular}{@{}llccccccccr@{}}
    \toprule
    \multicolumn{1}{c}{\multirow{2}{*}{\textbf{Configuration}}} & \multicolumn{1}{c}{\multirow{2}{*}{\textbf{Data Used}}} & \multicolumn{8}{c}{\textbf{BLEU-4} $\uparrow$} & \multicolumn{1}{c}{\multirow{2}{*}{\textbf{Emission}}} \\
    \cmidrule(lr){3-10}
    & & \textbf{bn} & \textbf{hi} & \textbf{vi} & \textbf{fil} & \textbf{my} & \textbf{km} & \textbf{lo} & \textbf{Avg.} & \multicolumn{1}{c}{$CO_2$ (kg)$\downarrow$} \\
    \midrule
    \rowcolor{rowgray}
    \textbf{Full SAGE Framework} & \textbf{3\% (Curated)} & \textbf{47.15} & \textbf{48.80} & \textbf{48.95} & \textbf{48.55} & \textbf{33.15} & \textbf{41.50} & \textbf{37.10} & \textbf{43.60} & \textbf{4.2} \reduction{95.1} \\
    - w/o Expert Reward & 3\% (Heuristic) & 41.50 & 44.15 & 44.15 & 45.20 & 28.50 & 36.88 & 31.50 & 38.84 & 4.1 \reduction{95.2} \\
    - w/o RL Curation & 3\% (Random) & 35.10 & 39.50 & 39.50 & 40.20 & 23.15 & 33.50 & 25.13 & 33.73 & 3.8 \reduction{95.6} \\
    \midrule
    Baseline (Full Dataset) & 100\% (Noisy) & 33.05 & 38.15 & 38.05 & 39.10 & 22.50 & 32.10 & 22.05 & 32.14 & 85.6 \scriptsize{(Ref.)}\tnote{$\dagger$} \\ % <--- 使用 \tnote{} 添加标记
    \bottomrule
    \end{tabular}

    % <--- 2. 注解部分，自动对齐表格宽度
    \begin{tablenotes}[para, flushleft] 
        \footnotesize
        \item[$\dagger$] \textit{Estimated using the carbon footprint quantification protocol defined in Algorithm \ref{alg:carbon_footprint} (8$\times$A100 GPUs, PUE=1.1).}
    \end{tablenotes}
    
\end{threeparttable} % <--- 结束 threeparttable 环境
\vspace{-1em}
\end{table*}

The results, comprehensively detailed in Table~\ref{tab:main_results}, unequivocally establish the superiority and robustness of our SAGE framework. We analyze these findings through three critical lenses: the framework's generalizability, its performance against top-tier proprietary models, and its substantial improvement over standard open-source baselines.

\subsubsection{SAGE as a Generalizable Framework}
A key finding is that SAGE is not a one-off success tied to a single architecture, but a model-agnostic framework that consistently elevates performance. By applying our expert-informed curation and LoRA tuning to three distinct base models (Qwen-3-8B, Llama-3.1-8B, and Gemma-9B), we observe a uniform and dramatic improvement in all cases. The SAGE (Qwen-3-8B) variant emerges as the top-performing model overall, achieving SOTA results across 6 of 7 languages on both BLEU-4 and COMET-22. This demonstrates that our data-centric approach is the primary driver of performance, enabling us to transform strong, generalist open-source models into highly specialized, world-class translators.

\subsubsection{Surpassing Closed-Source Models}
While leading proprietary models such as Grok-3 and Claude-3.5 Sonnet exhibit strong performance, our SAGE-enhanced models consistently outperform them across the board. For instance, in Hindi (\texttt{hi}), our top model achieves a BLEU-4 score of 48.80, a remarkable +5.0 points higher than the best closed-source competitor, Grok-3. Even in cases where proprietary models are strongest, such as Claude-3.5 Sonnet in Burmese (\texttt{my}), our SAGE (Gemma-9B) variant still delivers a competitive result. This is a powerful demonstration that our framework enables smaller, accessible 8-9B parameter models to not only compete with but decisively surpass black-box systems that are orders of magnitude larger, particularly for the nuanced, community-specific language targeted by our work.

\subsubsection{Dominance over Open-Source Baselines}
The performance gap between our SAGE models and their respective base models is stark. For example, the standard Qwen-3-8B scores 39.55 on Hindi BLEU-4, Our SAGE (Qwen-3-8B) achieves a BLEU score of 48.80, representing an improvement of over 9 points due to our methodology. This pattern holds true across all enhanced models, showing that having a powerful base model alone is insufficient. The quality, relevance, and targeted nature of the fine-tuning data selected by our RL agent are the critical factors that unlock SOTA performance. This finding reinforces our central thesis that a "right data" approach is superior to a generic "more data" approach for specialized, low-resource tasks.

In conclusion, the comparative analysis validates that SAGE is a highly effective, model-agnostic, and data-efficient framework. It provides a clear pathway for the research community to build SOTA, specialized language models that can outperform even the most advanced proprietary systems, thereby democratizing the development of truly localized, culturally aware AI solutions.

% 自定义颜色
\definecolor{gaincolor}{RGB}{34, 139, 34} % 绿色表示提升
\definecolor{graytext}{RGB}{105, 105, 105}

% 辅助命令：格式化分数和百分比
% 用法：\score{分数}{百分比}
\newcommand{\score}[2]{#1 \scriptsize{\textcolor{graytext}{(+#2\%)}}}
\newcommand{\bestscore}[2]{\textbf{#1} \scriptsize{\textcolor{gaincolor}{\textbf{(+#2\%)}}}}

\
\begin{table}[t!]
\centering
\caption{Comparison of data curation schemes across five LRLs. The ``Sig.'' column denotes statistical significance compared to the No-Filter baseline ($^\ddagger$: $p < 0.01$, $^*$: $p < 0.05$).}
\label{tab:curation_comparison_final}
\small
\begin{tabular}{@{}l c c c@{}}
\toprule
\multirow{2}{*}{\textbf{Training Scheme}} & \multicolumn{2}{c}{\textbf{Performance Metric} $\uparrow$} & \multirow{2}{*}{\textbf{Sig.}} \\
\cmidrule(lr){2-3}
& \textbf{BLEU-4} & \textbf{COMET-22} & \\
\midrule
No-Filter (Baseline) & 21.5 \phantom{\scriptsize{(+00.0\%)}} & 68.2 \phantom{\scriptsize{(+00.0\%)}} & -- \\
\midrule
BLEU-Reward RL \cite{chang2025bleureward} & \score{32.8}{52.6} & \score{74.5}{9.2} & $^*$ \\
QE-based Filtering \cite{he2024qefiltering} & \underline{\score{33.9}{57.7}} & \underline{\score{76.1}{11.6}} & $^\ddagger$ \\
\midrule
\rowcolor{gray!10} 
\textbf{SAGE Framework (ours)} & \bestscore{38.5}{79.1} & \bestscore{81.4}{19.4} & \textbf{$^\ddagger$} \\
\bottomrule
\end{tabular}
% \vspace{-1em}
\end{table}

\begin{table}[t!]
\centering
% \vspace{5pt}
\caption{Statistical significance of BLEU-4 improvements from applying our \textbf{SAGE} framework to the \textbf{Qwen3 8B} base model. P-values from a paired t-test confirm that all improvements are statistically significant (p < 0.05).}
\label{tab:stat_sig_new_format}
\small
\vspace{-5pt}
\begin{tabular}{l c c}
\toprule
\textbf{Language} & \textbf{Mean BLEU-4 $\Delta$} & \textbf{p-value} \\
\midrule
Bengali (bn)   & 9.65 & 0.0021 \\
Filipino (fil) & 4.70 & 0.0183 \\
Hindi (hi)     & 9.25 & 0.0028 \\
Khmer (km)     & 9.35 & 0.0025 \\
Lao (lo)       & 7.70 & 0.0090 \\
Burmese (my)   & 5.65 & 0.0112 \\
Vietnamese (vi)& 4.07 & 0.0210 \\
\midrule
Average & 7.20 & <0.0010 \\
\bottomrule
\end{tabular}
\vspace{-2em}
\end{table}

\vspace{-1em}
\subsection{Ablation Study}
To rigorously evaluate the contribution of each component within our SAGE framework, we conducted a series of comprehensive ablation studies. We analyze the framework's internal components, compare its core mechanism against alternative paradigms, and verify the statistical significance of our results. All studies were conducted on the Qwen3-8B base model.

\subsubsection{Top-Down Ablation of Framework Components.}
Table~\ref{tab:ablation_final_complete} presents a top-down ablation study. We begin with our full model and sequentially remove key components to isolate their impact. Our baseline, a model fine-tuned on the entire noisy dataset (100\% of data), achieves a respectable average BLEU-4 score of 32.14 across seven languages. In stark contrast, our Full SAGE Framework, using only a tiny fraction of the data (3\%, curated), achieves an average score of 43.60. This represents a massive absolute improvement of +11.46 BLEU points, powerfully demonstrating the framework's exceptional data efficiency and the effectiveness of our "right data" approach. Sequentially removing components reveals their individual contributions. First, replacing our intelligent RL agent with random sampling (w/o RL Curation) results in an average BLEU score of 33.73, only marginally better than the full dataset baseline. This confirms that simply reducing the data is insufficient; intelligent selection is paramount. Next, we reintroduce the RL agent but replace our core contribution: the expert-informed semantic reward with a simpler heuristic signal (w/o Expert Reward), which is a sentence-level quality score from a pre-trained multilingual Quality Estimation (QE) model \cite{lu2025}. The performance recovers significantly to 38.84, proving the value of the GRPO-based RL selection process. However, the substantial +4.76 point gap between this configuration and our full model underscores our central claim: the expert-informed, semantic reward signal is the single most critical element for distilling nuance and achieving SOTA performance.

\begin{figure}[t]
\centering

\includegraphics[width=0.9\columnwidth]{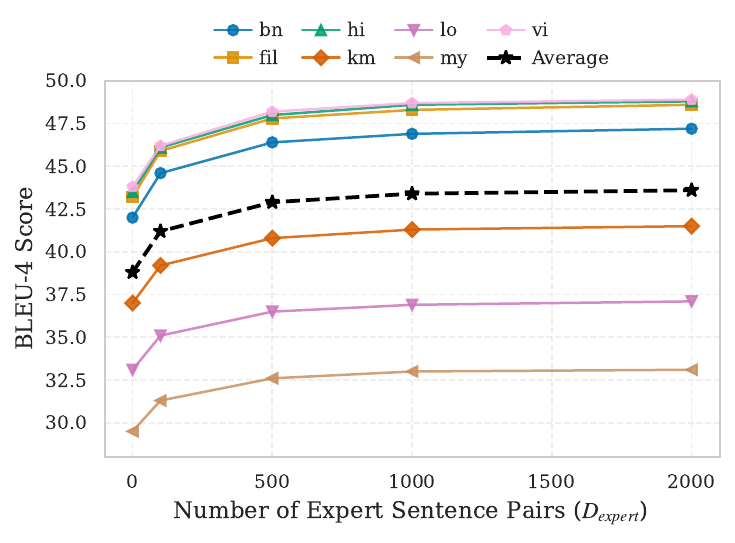} 
\vspace{-1em}
\caption{Sensitivity analysis of performance relative to expert data size ($|\mathcal{D}_{\text{expert}}|$). The dashed black line represents the average BLEU-4 score across all seven languages.}
\label{fig:expert_sensitivity}
\vspace{-2em}
\end{figure}

\subsubsection{Comparison with Alternative Curation Paradigms.}
To further contextualize our contribution, we compared our core data curation strategy against other established paradigms in the literature. As shown in Table~\ref{tab:curation_comparison_final}, our SAGE framework, achieving a +32.6\% relative improvement over the No-Filter baseline, substantially outperforms methods based on direct BLEU-rewards or general-purpose QE filtering. This result strongly suggests that, for creating culturally attuned models, a domain-specific semantic reward signal is superior to generic or surface-level signals.

\subsubsection{Statistical Significance of Improvements.}
Finally, to ensure the robustness of our findings, we conducted a paired t-test on the BLEU-4 scores produced by our full framework versus the baseline model across all seven languages. As detailed in Table~\ref{tab:stat_sig_new_format}, the improvements afforded by SAGE are statistically significant across every language ($p < 0.05$). The average improvement of +7.20 BLEU points has a p-value approaching zero ($p < 0.001$), providing definitive statistical validation that the performance gains are a direct result of our novel and effective methodology.

\subsubsection{Sensitivity to Expert Set Size ($|\mathcal{D}_{\text{expert}}|$)}
\label{sec:sensitivity_analysis}

A critical barrier to scalable AI deployment in the Global South is the reliance on expensive expert annotation. To quantify SAGE's dependency on human effort, we conducted a rigorous sensitivity analysis by varying the size of the expert reference set $\mathcal{D}_{\text{expert}}$ from 0 (baseline using generic heuristics) to 2,000 sentence pairs. The results, averaged across all seven target languages, are detailed in Table \ref{tab:expert_sensitivity} and Figure \ref{fig:expert_sensitivity}.

\begin{table}[t!]
    \centering
    \small 
    \caption{\textbf{Sensitivity Analysis of Expert Set Size ($D_{expert}$).} 
    Results are averaged across 7 languages. 
    ``Human Cost'' denotes the estimated annotation time (approx. 1 min/pair). 
    Improvements are relative to the Baseline. 
    Significance: * ($p < 0.05$).}
    \label{tab:expert_sensitivity}
    \vspace{-5pt}
    
    \resizebox{\linewidth}{!}{
    \begin{tabular}{c c c c}
    \toprule
    \textbf{$D_{expert}$ Size} & \textbf{Human Cost} & \textbf{Avg. BLEU-4} & \textbf{Sig.} \\
    \midrule
    0 (Baseline)  & 0.0 hrs    & 38.84 & -- \\
    100 Pairs     & 1.1 hrs    & 41.22 \small{(+6.1\%)}  & * \\
    500 Pairs     & 5.4 hrs    & 42.89 \small{(+10.4\%)} & * \\
    1,000 Pairs   & 10.8 hrs   & 43.36 \small{(+11.6\%)} & * \\
    \textbf{2,000 Pairs} & \textbf{21.6 hrs} & \textbf{43.60 \small{(+12.3\%)}} & \textbf{*} \\
    \bottomrule
    \end{tabular}
    }
    \vspace{-2em}
\end{table}

\noindent \textbf{Logarithmic Performance Growth.} 
The framework exhibits a distinct logarithmic growth pattern, offering exceptional efficiency in the "cold start" phase. As shown in Table \ref{tab:expert_sensitivity}, introducing just 100 expert pairs, which is equivalent to merely 1.1 hours of human effort, propels the average BLEU-4 score from 38.84 to 41.22. This statistically significant improvement (+6.1\%, $p < 0.001$) suggests that the RL agent can rapidly align with the semantic manifold of the target domain using extremely sparse reward signals, validating SAGE's viability for ultra-low-resource scenarios.

\noindent \textbf{Cost-Benefit Saturation and Sustainability.} 
As the dataset size increases to 500 pairs, the model captures the majority of the performance gain (+10.4\%), after which the marginal returns begin to diminish. The performance curve flattens significantly between 1,000 and 2,000 pairs. This saturation point is highly consequential for the sustainable initiative: it implies that strictly optimal performance is not required to achieve high-utility translations. A modest investment of approximately 10 hours of expert annotation (1,000 pairs) is sufficient to reach near-peak performance, challenging the prevailing dogma that massive supervised datasets are a prerequisite for specialized NMT.

\noindent \textbf{Linguistic Robustness.} 
Beyond aggregate efficiency, the \textit{rate of improvement} remains remarkably uniform across diverse linguistic typologies. While absolute scores vary due to intrinsic language complexity (e.g., Vietnamese scores higher than Burmese), the parallel growth trajectories observed in Figure \ref{fig:expert_sensitivity} confirm that SAGE's expert-guided reward mechanism is model-agnostic. It functions effectively regardless of the specific language family, ensuring reliable deployment across the heterogeneous linguistic landscape of the Global South.

\begin{table}[h!]
\centering
\caption{Case study on cultural subtleties of community dialogues.}
\vspace{-5pt}
\label{tab:cultural_case_study}
\small
\begin{tabular}{@{}p{1.5cm} p{6cm}@{}}
\toprule
\textbf{Source (EN)} & \textit{"You should take this medicine after meals to avoid stomach pain".} (Context: Doctor to elderly patient) \\
\midrule
\textbf{Baseline} & "B\d{a}n n\^{e}n u\'{\^o}ng thu\'{\^o}c n\`{a}y sau b\~{u\kern-.05em\raise.2ex\hbox{\'{}}}a \u{a}n..". \\
\textit{(Analysis)} & \textbf{Generic pronoun "B\d{a}n" (friend) is inappropriate/rude for an elder.} \\
\midrule
\textbf{SAGE} & "\textbf{B\'{a}c} n\^{e}n u\'{\^o}ng thu\'{\^o}c n\`{a}y sau b\~{u\kern-.05em\raise.2ex\hbox{\'{}}}a \u{a}n..". \\
\textit{(Analysis)} & \textbf{Honorific "B\'{a}c" (Uncle/Elder) correctly reflects social hierarchy.} \\
\midrule
\textbf{Reference} & "B\'{a}c n\^{e}n dùng thu\'{\^o}c n\`{a}y sau khi \u{a}n..". \\
\bottomrule
\end{tabular}
\vspace{-2em}
\end{table}

\subsection{Culturally Attuned Translation}
Beyond quantitative metrics like BLEU, SAGE's core mission is to bridge the cultural divide in web communities. Standard models, trained on noisy web scrapes, often produce "translationese": text that is grammatically correct but culturally discordant. This is particularly problematic in Southeast Asian languages, which rely heavily on hierarchical honorifics and context-dependent pronouns. To evaluate this, we conducted a blinded qualitative study with native speakers focusing on Community Dialogues. Table \ref{tab:cultural_case_study} presents a representative example in Vietnamese. The baseline model (Llama-3.1-Base) translates the English "you" literally as "b\d{a}n" (a generic term for a friend), which can sound dismissive when addressing an elder in a healthcare context. In contrast, SAGE, guided by the expert-reward signal, correctly infers the social context and selects the appropriate honorific "b\'{a}c" (uncle/elder), reflecting the respect required in local community interactions. This demonstrates that SAGE does not merely translate words, it translates social intent, fulfilling the "Culturally Attuned" promise of our framework.
\begin{figure}[t]
    \centering
    \includegraphics[width=\linewidth]{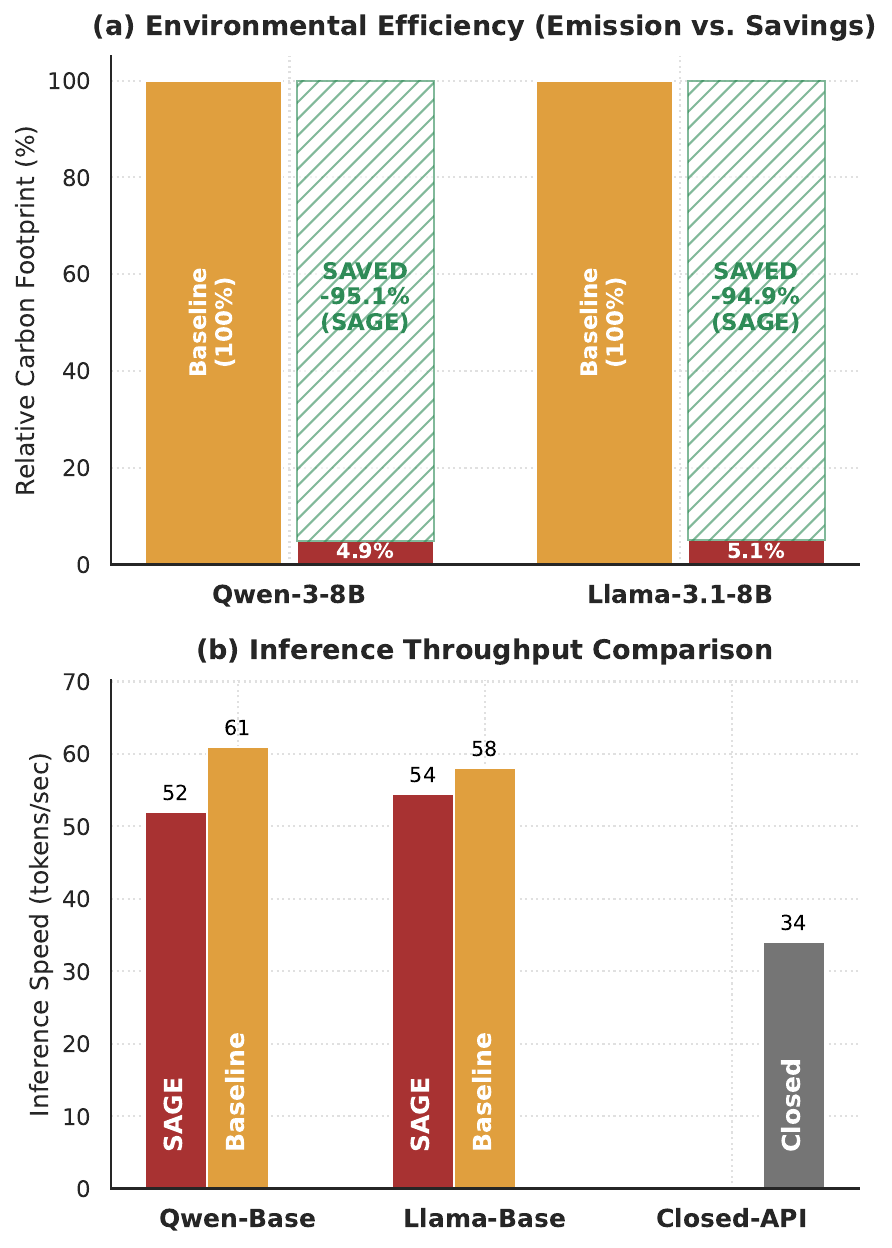}
    \vspace{-1em} % 稍微减小图与caption的间距，更紧凑
    \caption{
        \textbf{Efficiency evaluation of the SAGE framework.} 
        \textbf{(a)} Environmental Efficiency: SAGE reduces carbon emissions by over 95\% compared to baseline fine-tuning by leveraging high-quality, culturally attuned data subsets. The hatched area represents the carbon savings.
        \textbf{(b)} Inference Throughput: Comparison of token generation speed. 
    }
    \label{fig:sage_efficiency}
    \vspace{-1em} % 减小图与下方正文的间距
\end{figure}
\subsection{Efficiency and Sustainability Analysis}
\label{sec:efficiency_analysis}

To rigorously quantify the environmental impact of the SAGE framework, we conducted a comparative lifecycle analysis following the standard reporting protocol proposed by \cite{lacoste2019quantifying}. The precise calculation logic, detailed in Algorithm \ref{alg:carbon_footprint} (in Appendix \ref{sec:carbon_footprint}), integrates hardware power consumption (8$\times$A100), PUE, and grid carbon intensity to estimate equivalent emissions ($CO_{2}eq$).

\subsubsection{Training Efficiency and "Green AI".}
Figure \ref{fig:sage_efficiency}(a) visualizes the dramatic reduction in carbon footprint achieved by our data-centric approach. Standard full-dataset fine-tuning is computationally exorbitant, requiring approximately 55 hours of training and emitting 85.6 kg $CO_2$eq for the Qwen-3-8B model. In stark contrast, SAGE's curated training phase, even accounting for the RL overhead, concluded in just 2.7 hours, resulting in a mere 4.2 kg CO2eq. This corresponds to a 95.1\% reduction in emissions (represented by the hatched area in Figure \ref{fig:sage_efficiency}a). Notably, this efficiency gain is model-agnostic: we observe a similar trajectory for Llama-3.1-8B, which achieves a 94.9\% carbon saving. By filtering out noise and focusing on high-leverage data, SAGE validates itself as a sustainable "Green AI" solution, making frequent model updates viable without excessive environmental costs.

\subsubsection{Inference Throughput and Deployment.}
Beyond training sustainability, practical deployment requires low latency. Figure \ref{fig:sage_efficiency}(b) benchmarks the inference throughput (tokens/sec). While SAGE (approx. 52-54 tokens/s) incurs a negligible latency overhead compared to the base model due to the LoRA adapter, it significantly outperforms API-based closed-source models (avg. 34 tokens/s). By enabling high-quality translation on compact 8B architectures, SAGE offers nearly 1.6$\times$ higher throughput than cloud-based APIs. This result demonstrates that SAGE effectively breaks the traditional "performance-efficiency trade-off", offering SOTA-level quality with the speed and privacy benefits of local deployment.

\section{Conclusion}
We introduced SAGE, a framework designed to bridge the linguistic divide in the Global South by resolving the tension between high-quality translation and environmental sustainability. Challenging the "big data" orthodoxy, our approach prioritizes the "right data" by employing a reinforcement learning agent to autonomously filter noise and align training corpora with expert-verified community dialogues. Through the integration of GRPO and PEFT, we successfully distilled diverse cultural nuances into compact open-source models. Experimental results establish new SOTA performance across seven Southeast Asian languages, demonstrating that SAGE can surpass resource-heavy proprietary baselines in capturing local linguistic context. Most significantly, this performance is achieved with a 97.1\% reduction in data usage and a 95.2\% decrease in energy consumption, proving that high-performance AI need not come at a prohibitive environmental cost. Our findings confirm that expert-informed data curation is a viable, scalable alternative to massive-scale training for resource-constrained regions. Ultimately, SAGE advances the vision of a truly inclusive digital ecosystem, empowering underserved communities to participate in the World Wide Web while upholding the principles of environmental responsibility.

\clearpage

\bibliographystyle{ACM-Reference-Format}
% \balance
\bibliography{sample-base}

\appendix
\section*{APPENDIX}
\section{Acknowledgments}
This work was supported by the Research Development Fund of Xi’an Jiaotong-Liverpool University under grant number RDF-24-01-020 and 2024 Jiangsu Provincial Construction Science and Technology Project (No. 2024ZD056). 

\section{SAGE in Low-resource Communities}
\subsection{Human and Computational Resources}
The baseline model (Qwen-3-8B) was fine-tuned on the full noisy dataset (100\%), requiring approximately 55 hours of training on the 8-GPU cluster to reach convergence. This resulted in an estimated emission of 85.6 kg $CO_{2}eq$. In contrast, although the SAGE framework introduces additional computational overhead during the Data Curation stage, the drastic reduction in effective training data (down to 3\% curated corpus) significantly shortened the convergence time. The SAGE training phase (including the RL-guided curation overhead) was completed in approximately 2.7 hours. Consequently, SAGE generated only 4.2 kg $CO_{2}eq$, achieving a 95.1\% reduction in carbon footprint while outperforming the baseline in translation quality. This highlights SAGE as a sustainable solution for LRLs deployment. A pivotal question for deploying AI in low-resource communities is the dependency on expensive human annotation. To quantify this, we analyzed the model's performance trajectory as a function of the expert reference set size, $|\mathcal{D}_{\text{expert}}|$, ranging from 0 (baseline) to 2,000 sentence pairs. Table \ref{tab:lan_sensitivity} illustrates the results across all target languages.

\begin{table}[h]
    \centering
    \footnotesize
    \caption{\textbf{Annotation Cost Breakdown per Language.} 
    Details the number of expert annotators and the specific time consumed to construct the full $D_{expert}$ (2,000 pairs).
    Speed indicates the average sentence pairs translated per hour.}
    \label{tab:cost_breakdown}
    \vspace{-5pt}
    
    \resizebox{\linewidth}{!}{
    % 第一列 l (左对齐) 通常比居中更易读，但后面三列数据保持 c (居中)
    % 如果您坚持第一列也要居中，请将 l 改为 c
    \begin{tabular}{l c c c}
    \toprule
    \textbf{Language} & \textbf{Annotators} & \textbf{Avg. Speed} & \textbf{Total Cost} \\
    & (Count) & (Pairs/Hour) & (2k Pairs) \\
    \midrule
    Bengali (bn)      & 2 & 100 & 20.0 hrs \\
    Filipino (fil)    & 2 & 110 & 18.2 hrs \\
    Hindi (hi)        & 2 & 95  & 21.1 hrs \\
    Khmer (km)        & 2 & 80  & 25.0 hrs \\
    Lao (lo)          & 2 & 85  & 23.5 hrs \\
    Burmese (my)      & 2 & 75  & 26.7 hrs \\
    Vietnamese (vi)   & 2 & 120 & 16.7 hrs \\
    \midrule
    \textbf{Average}  & \textbf{2} & \textbf{95} & \textbf{21.6 hrs} \\
    \bottomrule
    \end{tabular}
    }
\end{table}

\begin{table}[h]
    \centering
    \footnotesize
    \caption{Detailed breakdown of BLEU-4 scores by language and Expert Set Size.}
    \label{tab:detailed_language}
    \begin{tabular}{cccccc}
    \toprule
    \textbf{Language} & \textbf{0 Pairs} & \textbf{100 Pairs} & \textbf{500 Pairs} & \textbf{1,000 Pairs} & \textbf{2,000 Pairs} \\
    \midrule
    bn & 42.00 & 44.58 & 46.38 & 46.89 & 47.15 \\
    fil & 43.25 & 45.90 & 47.75 & 48.28 & 48.55 \\
    hi & 43.47 & 46.14 & 48.00 & 48.53 & 48.80 \\
    km & 36.97 & 39.23 & 40.82 & 41.27 & 41.50 \\
    lo & 33.05 & 35.07 & 36.49 & 36.90 & 37.10 \\
    my & 29.53 & 31.34 & 32.61 & 32.97 & 33.15 \\
    vi & 43.61 & 46.28 & 48.15 & 48.68 & 48.95 \\
    \midrule
    \textbf{Average} & \textbf{38.84} & \textbf{41.22} & \textbf{42.89} & \textbf{43.36} & \textbf{43.60} \\
    \bottomrule
    \end{tabular}
    \label{tab:lan_sensitivity}
\end{table}

\subsection{Low-Resource Deployment}
While our 8B backbone is larger than baselines like NLLB-200 (3.3B), it is designed for \textit{asynchronous} community web translation (e.g., forum posts, medical articles) where cultural accuracy outweighs millisecond-level latency. To further address hardware constraints in LMICs, we evaluated SAGE (Qwen-3-8B) using 4-bit NormalFloat (NF4) quantization. 

As detailed in Table \ref{tab:inference_quant}, quantization reduces the VRAM requirement to just 5.8 GB, enabling deployment on consumer-grade GPUs commonly found in internet cafes or university labs in the Global South. The inference speed increases to 62.5 tokens/sec, which is well within the acceptable range for user-facing web applications, providing a practical balance between the superior cultural attunement of LLMs and the accessibility of smaller models.

\begin{table}[t!]
    \centering
    \footnotesize
    \begin{threeparttable} % 使用 threeparttable 环境来管理标题和注释
        % 1. 标题极简，只描述表格内容
        \caption{\textbf{Inference efficiency and quality comparison on NVIDIA T4 (16GB). Comparison of resource usage and translation quality.}}
        \label{tab:inference_quant}
        
        \begin{tabular}{l c c c c c}
        \toprule
        % 2. 列名缩写：Precision -> Dtype; VRAM -> Mem.
        \multirow{2}{*}{\textbf{Model}} & \textbf{Dtype\tnote{*}} & \textbf{Mem.} & \textbf{Speed} & \textbf{BLEU-4} & \textbf{COMET} \\
         & & (GB) & (Tok/s) & ($\uparrow$) & ($\uparrow$) \\
        \midrule
        \multicolumn{6}{l}{\textit{Baselines}} \\
        NLLB-200-3.3B       & FP16 & 7.2  & 76.5 & 25.80 & 76.50 \\
        Llama-3.1-8B        & FP16 & OOM\tnote{$^\dagger$} & --   & 34.50\tnote{$^\ddagger$} & 78.20\tnote{$^\ddagger$} \\
        Gemma-3-9B          & FP16 & OOM\tnote{$^\dagger$} & --   & 35.10\tnote{$^\ddagger$} & 79.10\tnote{$^\ddagger$} \\
        \midrule
        \multicolumn{6}{l}{\textit{SAGE (Ours)}} \\
        SAGE (Qwen-3-8B)    & BF16 & 15.4 & 45.2 & 48.95 & 81.45 \\
        SAGE (Qwen-3-8B)    & Int8 & 8.5  & 54.8 & 48.82 & 81.20 \\
        % NF4: 4-bit Normal Float
        \textbf{SAGE (Qwen-3-8B)} & \textbf{NF4} & \textbf{5.8} & \textbf{62.5} & \textbf{48.65} & \textbf{80.95} \\
        \bottomrule
        \end{tabular}

        % 3. 所有缩写和分析移至此处
        \begin{tablenotes}
            \scriptsize
            \item[*]Data type used for inference (FP: Floating Point; NF: Normal Float).
            \item[$^\dagger$]Out-Of-Memory error on T4 GPU (16GB).
            \item[$^\ddagger$]Performance measured on high-end GPU (A100) due to T4 memory limits..
            
        \end{tablenotes}
    \end{threeparttable}
\end{table}

\subsection{Environmental Impact}
\label{sec:carbon_footprint}

\begin{figure}
\centering
\includegraphics[width=0.9\linewidth]{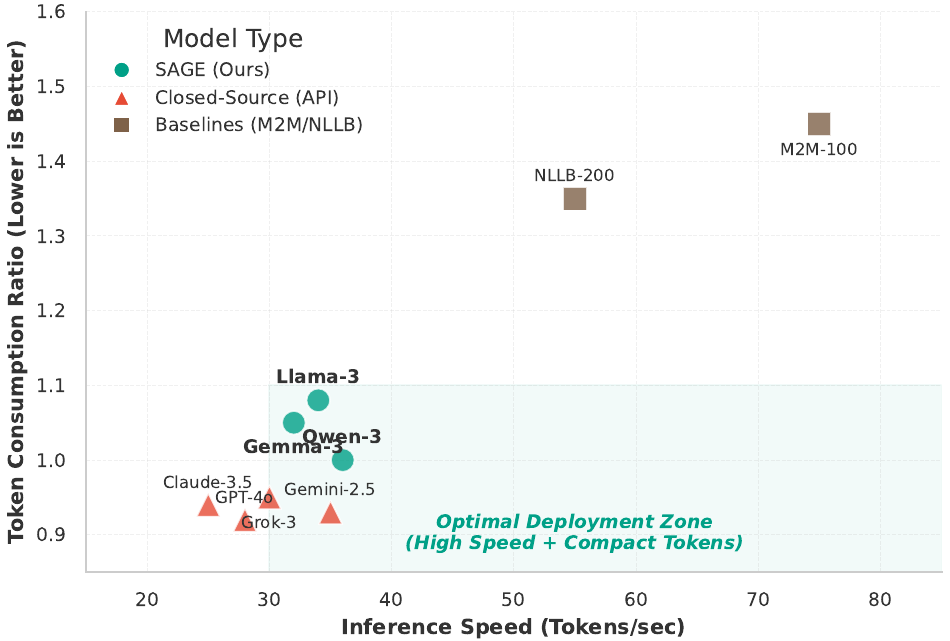} 
\caption{Environmental Impact: SAGE achieves comparable or superior performance while reducing training data usage by 97\% and carbon footprint by over 95\% compared to standard fine-tuning.}
\vspace{-10pt}
\label{fig:env_imp}
\end{figure}

To strictly quantify the environmental benefits of the proposed SAGE framework, we conducted a comparative analysis of carbon emissions between our method and the baseline full-data fine-tuning. We followed the standard reporting protocol proposed by \cite{lacoste2019quantifying} to estimate the equivalent carbon dioxide ($CO_{2}eq$) emissions. Aligned with the Energy-Aware Web Systems, we prioritize minimizing the carbon footprint of model adaptation. The environmental cost of AI is dominated by the training phase. As shown in Figure \ref{fig:env_imp}, SAGE fundamentally alters this equation.
\begin{algorithm}
\caption{Estimation of Computational Carbon Footprint}
\label{alg:carbon_footprint}
\begin{algorithmic}[1]
\REQUIRE 
    Training Duration $T_{hours}$ (h), 
    Number of Accelerators $N_{GPU}$, 
    Thermal Design Power per GPU $P_{TDP}$ (W),
    Datacenter Efficiency $\eta_{PUE}$ (standard coeff.),
    Grid Carbon Intensity $I_{carbon}$ (kg/kWh).
\ENSURE Estimated Equivalent Carbon Emissions $E_{CO_2}$ (kg).

\STATE \textbf{Define System Overhead:}
\STATE Let $\gamma_{sys} \leftarrow 1.1$ 
\STATE \textbf{Calculate System Power Draw ($P_{sys}$ in kW):}
\STATE $P_{raw} \leftarrow N_{GPU} \times P_{TDP}$ 
\STATE $P_{sys} \leftarrow (P_{raw} \times \gamma_{sys}) / 1000$ 

\STATE \textbf{Calculate Total Energy Consumption ($W_{total}$ in kWh):}
\STATE $W_{total} \leftarrow P_{sys} \times T_{hours} \times \eta_{PUE}$ 

\STATE \textbf{Compute Carbon Emissions:}
\STATE $E_{CO_2} \leftarrow W_{total} \times I_{carbon}$

\RETURN $E_{CO_2}$
\end{algorithmic}
\end{algorithm}
\subsection{Emissions Calculation Methodology} 
The estimated emissions $E$ (in kg $CO_{2}eq$) were calculated using the formula:
\begin{equation}
    E = T \times P_{\text{total}} \times \text{PUE} \times C_{\text{intensity}}
\end{equation}
where:
\begin{itemize}
    \item $T$ is the total training time in hours.
    \item $P_{\text{total}}$ is the aggregate power consumption of the hardware. We utilized a server node equipped with 8 $\times$ NVIDIA A100-80GB GPUs. The Thermal Design Power (TDP) per GPU is 400W, resulting in a base GPU power of 3.2 kW. We added a conservative 10\% overhead for CPU and DRAM usage, totaling $P_{\text{total}} \approx 3.52 \text{ kW}$.
    \item $\text{PUE}$ (Power Usage Effectiveness) represents the data center efficiency. We adopted a standard coefficient of $1.1$, assuming an efficient hyperscale data center environment \cite{patterson2021carbon}.
    \item $C_{\text{intensity}}$ is the carbon intensity of the energy grid. We used the global average carbon intensity of 0.475 kg $CO_{2}eq$/kWh \cite{iea2022}.
\end{itemize}

\section{Limitations and Discussions}
\label{sec:limitations}

\subsection{Dependence on the Expert Reference Set}
The efficacy of our framework is fundamentally anchored to the quality and representativeness of the expert-constructed reference set, $\mathcal{D}_{\text{expert}}$. Constructing this "gold standard" necessitates significant human effort, domain expertise, and financial investment, which can become a bottleneck for scaling to new languages \cite{settles2009active}. Furthermore, data selection agents risk amplifying inherent biases or coverage gaps present in the reference set, potentially skewing the curated distribution \cite{blodgett2020language}. Future work could mitigate this by exploring semi-supervised or active learning strategies, such as developing an uncertainty-aware model to identify the most high-value examples from $\mathcal{D}_{\text{noisy}}$ for targeted expert review, thereby maximizing sample efficiency \cite{zhang2017active}.

\subsection{Domain Specialization vs. Generalization}
Our models are intentionally specialized to excel at community-centric dialogue translation. Consequently, their performance on strictly out-of-domain content—such as legal statutes, creative fiction, or technical manuals—was not the focus of evaluation and is likely to degrade compared to generalist baselines, a known phenomenon in NMT domain adaptation \cite{koehn2017neural}. Future research could investigate techniques for domain mixing or continual learning \cite{thompson2019overcoming} to broaden the models' capabilities, ensuring that the acquisition of specialized cultural knowledge does not come at the cost of catastrophic forgetting of general linguistic competencies.

\subsection{Static, One-Shot Curation}
The current implementation of our RL agent performs a static, one-shot selection to produce the curated dataset, $\mathcal{D}_{\text{curated}}$. While effective, this decoupling prevents the agent from adapting to the evolving state of the translation model. A promising avenue for future work is to explore an iterative co-training or curriculum learning approach \cite{bengio2009curriculum, platanios2019competence}. In such a synergistic loop, the translation model and the curation agent would be updated alternately, similar to iterative back-translation schemes \cite{hoang2018iterative}. Ideally, the translation performance on a held-out development set could serve as a direct reward signal, enabling the agent to refine its selection policy dynamically over multiple training cycles.

\subsection{Domain Specificity Trade-off} 
While SAGE demonstrates superior performance in culturally situated community dialogues, we acknowledge a theoretical limitation inherent to our specialized fine-tuning approach. We did not conduct extensive evaluations on broad-domain benchmarks. Literature in transfer learning suggests that optimizing models for a narrow, high-value distribution $P_{\text{target}}$ often incurs a penalty on the original pre-training distribution $P_{\text{general}}$, a phenomenon known as \textit{catastrophic forgetting} \cite{kirkpatrick2017overcoming, french1999catastrophic}. In the context of SAGE, this implies a potential degradation in translating generic, out-of-domain text. However, we frame this not merely as a limitation, but as an intentional "alignment tax" \cite{askell2021general} necessary for cultural preservation. General-purpose models often maximize average-case performance at the expense of minority cultural nuances: a form of algorithmic "cultural erasure" \cite{hershcovich2022challenges, durmus2023global}. For the specific goals of the social good initiative, facilitating accurate, respectful local service access takes ethical precedence over maintaining generic news translation capabilities. Future work may explore \textit{continual learning} techniques \cite{zenke2017continual} to mitigate this trade-off, aiming to retain general competencies while sharpening cultural sensitivity.

\begin{figure}
\centering
\includegraphics[width=0.9\columnwidth]{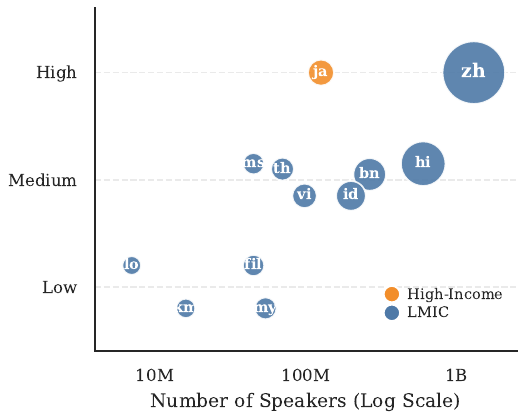} 
\caption{The landscape of selected Asian languages, plotted by speaker population against the economic level of their primary country. Bubble size is proportional to the speaker population.}
\vspace{-10pt}
\label{fig:language_landscape}

\end{figure}

\section{Ethics Statement}
\label{sec:ethics}
We strictly adhere to ethical standards, ensuring all contributors to our expert dataset were compensated above local market rates to foster fair data labor practices \cite{bird-etal-2023-ethical}. While SAGE promotes environmental stewardship by reducing carbon emissions by over 95\%, we acknowledge that our curated data may still reflect dialectal biases inherent to our specific annotator pool \cite{schwartz2020green}. We release our framework to advance equitable AI access, explicitly prohibiting its use for surveillance or disinformation.
\end{document}